\documentclass[sigconf,nonacm]{acmart} %anonymous
\usepackage[stable]{footmisc}
\AtBeginDocument{%
  \providecommand\BibTeX{{%
    \normalfont B\kern-0.5em{\scshape i\kern-0.25em b}\kern-0.8em\TeX}}}

\setcopyright{rightsretained}
\copyrightyear{2023}

\begin{document}

%%
%% The "title" command has an optional parameter,
%% allowing the author to define a "short title" to be used in page headers.
\title{GRASP: Accelerating Shortest Path Attacks via Graph Attention}
\titlenote {Distribution Statement A. Approved for public release. Distribution is unlimited. This material is based upon work supported by the Under Secretary of Defense for Research and Engineering under Air Force Contract No. FA8702-15-D-0001. Any opinions, findings, conclusions or recommendations expressed in this material are those of the author(s) and do not necessarily reflect the views of the Under Secretary of Defense for Research and Engineering. © 2023 Massachusetts Institute of Technology. Delivered to the U.S. Government with Unlimited Rights, as defined in DFARS Part 252.227-7013 or 7014 (Feb 2014). Notwithstanding any copyright notice, U.S. Government rights in this work are defined by DFARS 252.227-7013 or DFARS 252.227-7014 as detailed above. Use of this work other than as specifically authorized by the U.S. Government may violate any copyrights that exist in this work.}

% DEGAS: Deep learning Enhanced Graph Attention for shortest path attack Strategies
% SPRINT: Shortest Path attack using Relevance Identification and deep Neural neTworks
% GRASS: Graph Attention-based Subgraph Selection for shortest path attacks

%%
%% The "author" command and its associated commands are used to define
%% the authors and their affiliations.
%% Of note is the shared affiliation of the first two authors, and the
%% "authornote" and "authornotemark" commands
%% used to denote shared contribution to the research.
\author{Zohair Shafi}
\email{shafi.z@northeastern.edu}
\affiliation{
    Northeastern University
    \city{Boston}
    \state{MA}
    \country {USA}
}

\author{Benjamin A. Miller}
\email{miller.be@northeastern.edu}
\affiliation{
    Northeastern University
    \city{Boston}
    \state{MA}
    \country {USA}
}

\author{Ayan Chatterjee}
\email{chatterjee.ay@northeastern.edu}
\affiliation{
    Northeastern University
    \city{Boston}
    \state{MA}
    \country {USA}
}

\author{Tina Eliassi-Rad}
\email{t.eliassirad@northeastern.edu}
\affiliation{
    Northeastern University
    \city{Boston}
    \state{MA}
    \country {USA}
}

\author{Rajmonda S. Caceres}
\email{rajmonda.caceres@ll.mit.edu}
\affiliation{
    MIT Lincoln Laboratory
    \city{Lexington}
    \state{MA}
    \country {USA}
}

%%
%% By default, the full list of authors will be used in the page
%% headers. Often, this list is too long, and will overlap
%% other information printed in the page headers. This command allows
%% the author to define a more concise list
%% of authors' names for this purpose.
\renewcommand{\shortauthors}{Shafi et al.}

%%
%% The abstract is a short summary of the work to be presented in the
%% article.

\begin{abstract}
Recent advances in machine learning (ML) have shown promise in aiding and accelerating classical combinatorial optimization algorithms. ML-based speed ups that aim to learn in an end to end manner (i.e., directly output the solution) tend to trade off run time with solution quality. Therefore, solutions that are able to accelerate existing solvers while maintaining their performance guarantees, are of great interest.  We consider an APX-hard problem, where an adversary aims to attack shortest paths in a graph by removing the minimum number of edges. We propose the \textbf{GRASP} algorithm: \textbf{Gr}aph \textbf{A}ttention \textbf{A}ccelerated \textbf{S}hortest \textbf{P}ath Attack, an ML aided optimization algorithm that achieves run times up to 10x faster, while maintaining the quality of solution generated. GRASP uses a graph attention network to identify a smaller subgraph containing the combinatorial solution, thus effectively reducing the input problem size. Additionally, we demonstrate how careful representation of the input graph, including node features that correlate well with the optimization task, can highlight important structure in the optimization solution.

%investigate the role of  role of node features in the effectiveness of our approach. 
\end{abstract}


%% A "teaser" image appears between the author and affiliation
%% information and the body of the document, and typically spans the
%% page.

% \received{20 February 2007}
% \received[revised]{12 March 2009}
% \received[accepted]{5 June 2009}

%%
%% This command processes the author and affiliation and title
%% information and builds the first part of the formatted document.
\maketitle

\section{Introduction}
ML algorithms have had great success in learning a variety of inference tasks, and more recently, they have been shown to accelerate and improve classically difficult combinatorial optimization problems. Following Bengio et al's survey~\cite{bengio2021machine}, we observe that there are essentially two categories for ML-based combinatorial optimization: one that uses ML in an end to end fashion, and another that uses ML to aid classical algorithms. Recent work by  Veli{\v{c}}kovi{\'c} et al. \cite{velivckovic2022clrs},\cite{velivckovic2021neural} falls in the first category where the authors make progress by aligning the ML model to the structure of an algorithm---for example, dynamic programming---and employing imitation learning to mimic each step of the algorithm. Gasse et al. \cite{gasse2019exact} fall into the second category of using ML to learn offline, expensive to compute heuristics that are used in MIP solvers.  
%like sorting, searching and dynamic programming. 
%graph algorithms, string algorithms and geometric algorithms. 
Our contributions focus on the second category -- i.e., ML aided approaches to classical algorithms -- since the form of alignment discussed in end to end solutions are not straightforward for problems with more complex components. We consider one such algorithm by Miller et al., PATHATTACK \cite{miller2022attacking}, where an adversary wants to force traffic from a pair of nodes to take a particular route by removing a minimum number of edges in the graph. %The problem setting is NP-complete, but can be recast as an instance of the Weighted Set Cover problem. A brief overview of the PATHATTACK algorithm is shown in Figure \ref{fig:pathattack}. The algorithm has two main components; a constraint generation routine and a optimization component that utilizes a traditional combinatorial solver. Each component is an algorithm in its own and subsequent components rely on the results of the previous component being accurate.
Minor approximation errors can accumulate along components of the algorithm, leading to sub-optimal solutions. Our goal is to speed up the PATHATTACK algorithm while ensuring that the quality of solutions does not decrease. To achieve this goal, we examine alternate ways of incorporating ML into combinatorial optimization. To that end, we propose the \textbf{GRASP} algorithm: \textbf{Gr}aph \textbf{A}ttention \textbf{A}ccelerated \textbf{S}hortest \textbf{P}ath Attack, that achieves run times up to 10x faster, while maintaining the quality of solution generated by PATHATTACK. GRASP uses a graph attention network to identify a smaller subgraph containing the combinatorial solution, thus effectively reducing the input problem size. We demonstrate the speed up achieved by GRASP across a variety of synthetic and real-world graphs with varying topologies, graph characteristics and sizes. We also discuss the impact of node features on various topologies.

\section{Problem Definition and Methods}
\begin{figure*}
    \centering
    \includegraphics[width=1.0\linewidth]{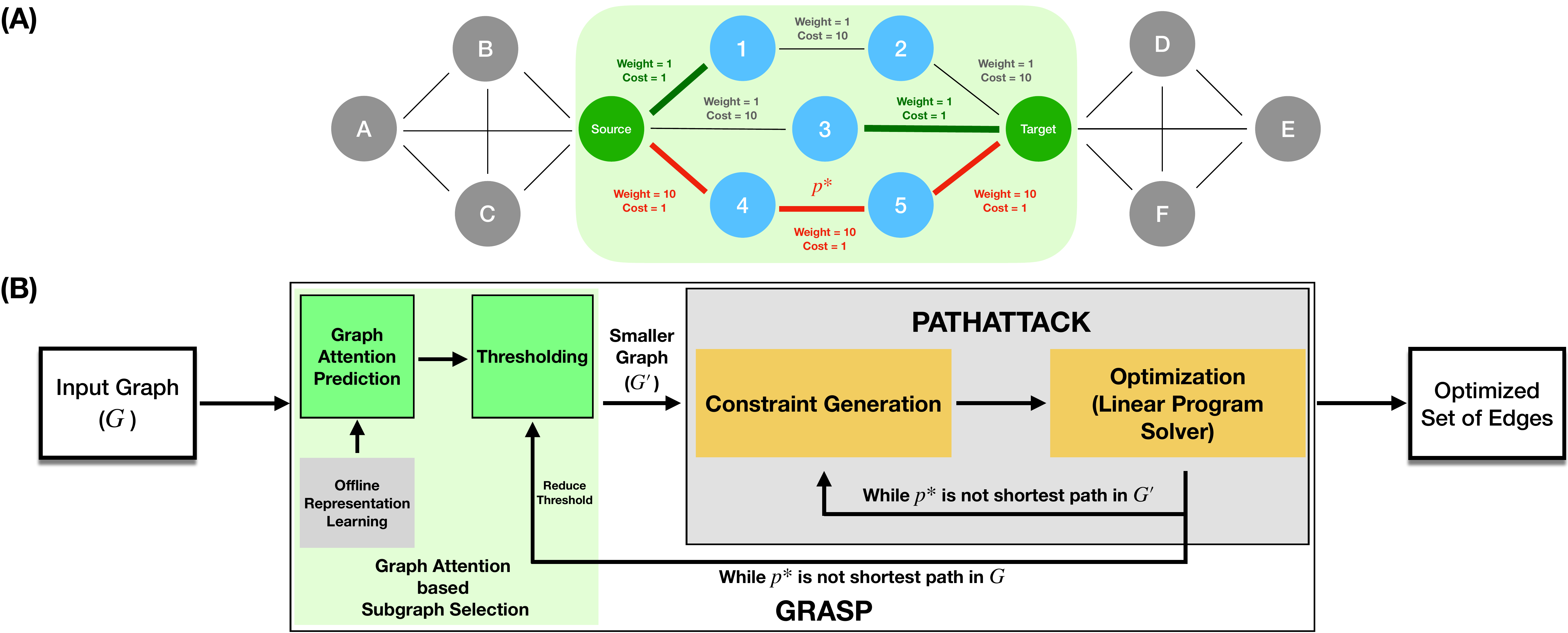}
    \caption{An overview of PATHATTACK. (A) The Force Path Cut problem: Given a source and target, we aim to make $p^*$ (edges in red) the shortest path by deleting the minimum cost of edge set (edges in green). (B) Our proposed solution, GRASP, identifies the subgraph (highlighted in green) that contains all paths of interest, i.e., paths between source and target that are shorter than $p^*$ and passes it to PATHATTACK. It can iteratively increase the subgraph size by reducing the threshold until $p^*$ is the shortest path returned by PATHATTACK. Observe how nodes in grey do not contribute to the solution, but add to the problem size.}
    \label{fig:pathattack}
\end{figure*}
PATHATTACK~\cite{miller2022attacking} provides an approximate solution to \emph{the Force Path Cut problem}, where the objective is to find an edge set within a budget constraint whose removal makes a particular path the shortest. This problem is APX-hard. However, it can be recast as an instance of the Weighted Set Cover problem, which enables an approximation algorithm with a logarithmic approximation factor in the worst case. A brief overview of the PATHATTACK algorithm is shown in Figure \ref{fig:pathattack}. The algorithm has two main components: a constraint generation routine and a optimization component that utilizes a traditional linear solver. %Each component is an algorithm in its own and subsequent components rely on the results of the previous component being accurate.
The size of the linear constraint matrix used by PATHATTACK is $(2|E|+|P|+1)\times |E|$, where $P$ is the set of constraint paths identified by the constraint generation procedure and $E$ is number of edges. Thus, considering a subset of the graph---and thus a much smaller edge set---enables a substantial speedup of the algorithm. 

Our proposed approach, GRASP, uses a learned model to help reduce the input problem size -- i.e., given a large graph, GRASP finds a subgraph that contains all paths relevant to the task of attacking the shortest paths between a given source and target. Such a subgraph is highlighted in Figure \ref{fig:pathattack}(A) in green, where nodes in grey are of no impact to the problem, but add additional complexity. This subgraph is passed into the PATHATTACK algorithm to arrive at a solution (\ref{fig:pathattack}(B)), thereby ensuring that the final output does not compromise on the quality of the solution, while achieving a speed up in run time. If the sampled graph does not yield a valid solution (i.e., $p^*$ is not the shortest path), this is because a path that competes with $p^*$ to be shortest was removed by the subgraph selection procedure. We then reduce the selection threshold and run PATHATTACK with a larger subgraph.

%\subsection{Embedding Visualization}
\subsection{Graph Embeddings Highlight Optimization Solution Structure }
\label{viz}

The field of Graph Representation Learning has demonstrated the important role of learned features in improving many inference tasks on graphs. We explore if learned features can similarly reveal important structure about the PATHATTACK solution space. In particular we are interested to understand if edges that fall in the optimal solution are clustered or easy to separate in some learned feature space. This type of structure can then be utilized to reduce the dimensionality of the optimization space.

%To learn a model to predict a subgraph, we first investigate the signal present in embeddings when using an unsupervised embedding algorithm. 
We first explore unsupervised representation learning of edges in the original graph using the Deep Graph Infomax (DGI) \cite{velickovic2019deep} algorithm. We embed a variety of synthetic graphs ($\sim$1000 nodes) into low-dimensional space (number of dimensions used in our experiments is 32)
%followed by using TSNE \cite{van2008visualizing} to visualize them in two dimensions. 
and generate edge embeddings by averaging its node embeddings. We also incorporate additional path information into the embeddings that we hypothesize might play a role in highlighting any structure relevant to the optimization task. More specifically, we use the following three sets of features: 
\begin{itemize}
    \item Structural Features - following Ghasemian et al. \cite{ghasemian2020stacking}, these include local and global features such as Degree, Clustering Coefficient, Katz Centrality, Page Rank, Eigenvector Centrality, Structural Holes Constraint, Average Neighbor Clustering, number of Ego Net edges and Node Betweenness. 
    \item Flow between Source and Target nodes, i.e., we compute the max flow between the source and target nodes using the weights on edges as the capacity. The flow value on each edge is then used as a feature. 
    \item Personalized Page Rank (PPR) vectors for each node along the desired path ($p^*$). Observe that this gives us a node feature vector for each node, which is of the length of the number of edges of $p^*$. More concretely, if $p^*$ has 5 edges, then every node in the graph gets a node feature vector of dimensions $\mathbb{R}^{5 \times 1}$. We pad these vectors up to a maximum length of 64 to account for varying lengths of $p^*$. 
\end{itemize}

\begin{figure}
    \centering
    \includegraphics[width=0.99\linewidth]{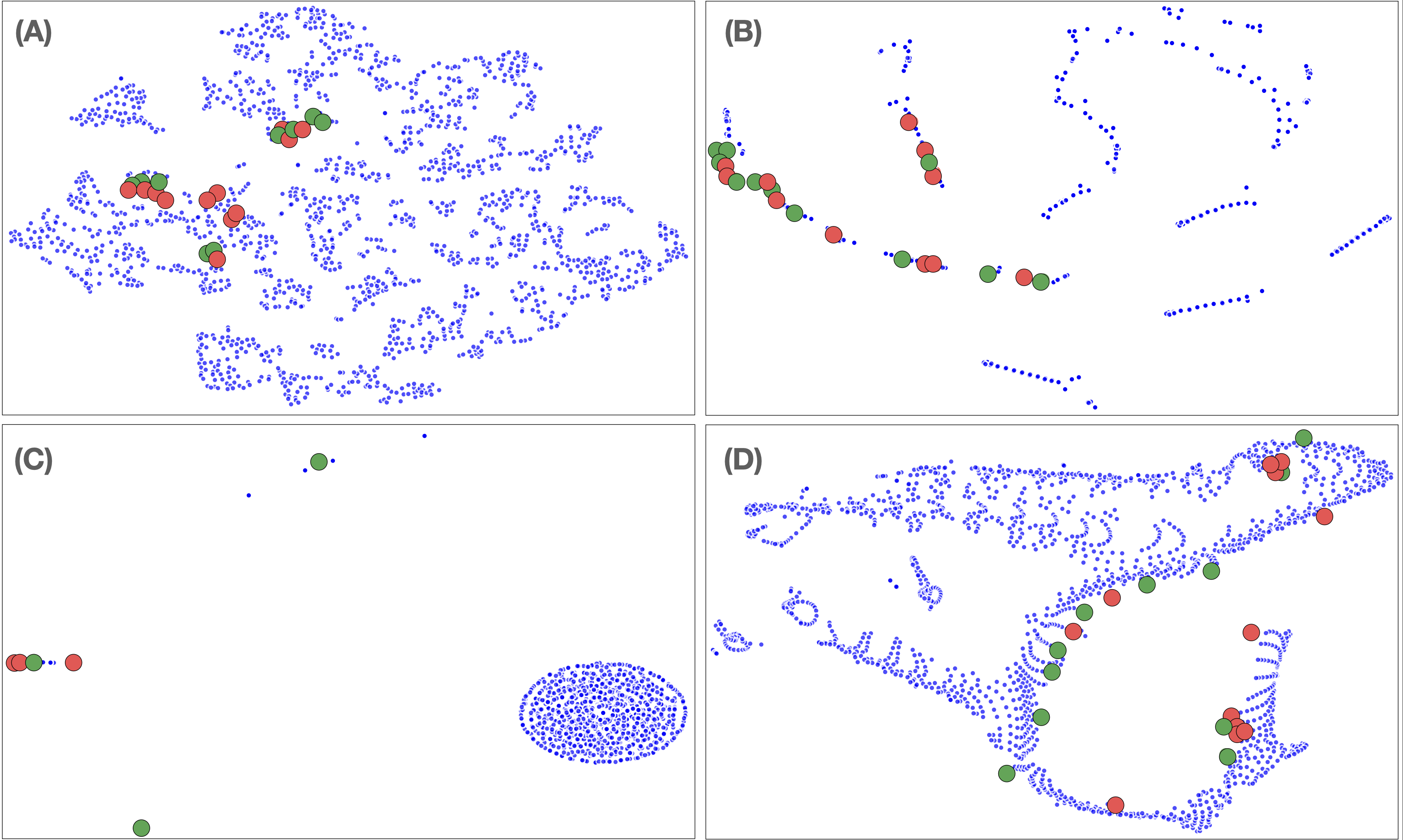}
    \caption{Visualizing edge embeddings for a 900-node lattice graph (30 $\times$ 30). Shown in red are edges that PATHATTACK deleted, shown in green are edges that are part of $p^*$, with blue being all other edges. The graph was embedded into 32 dimensions using DGI. Node features used were (A) no node features (B) structural features (C) flow values (D) Personalized Page Rank vectors. Observe that for the lattice graph, using flow values as node features (C) leads to the deleted edges (red) separating out from the rest of the edges.}
    \label{fig:lattice_dgi}
\end{figure}

\begin{figure}
    \centering
    \includegraphics[width=0.99\linewidth]{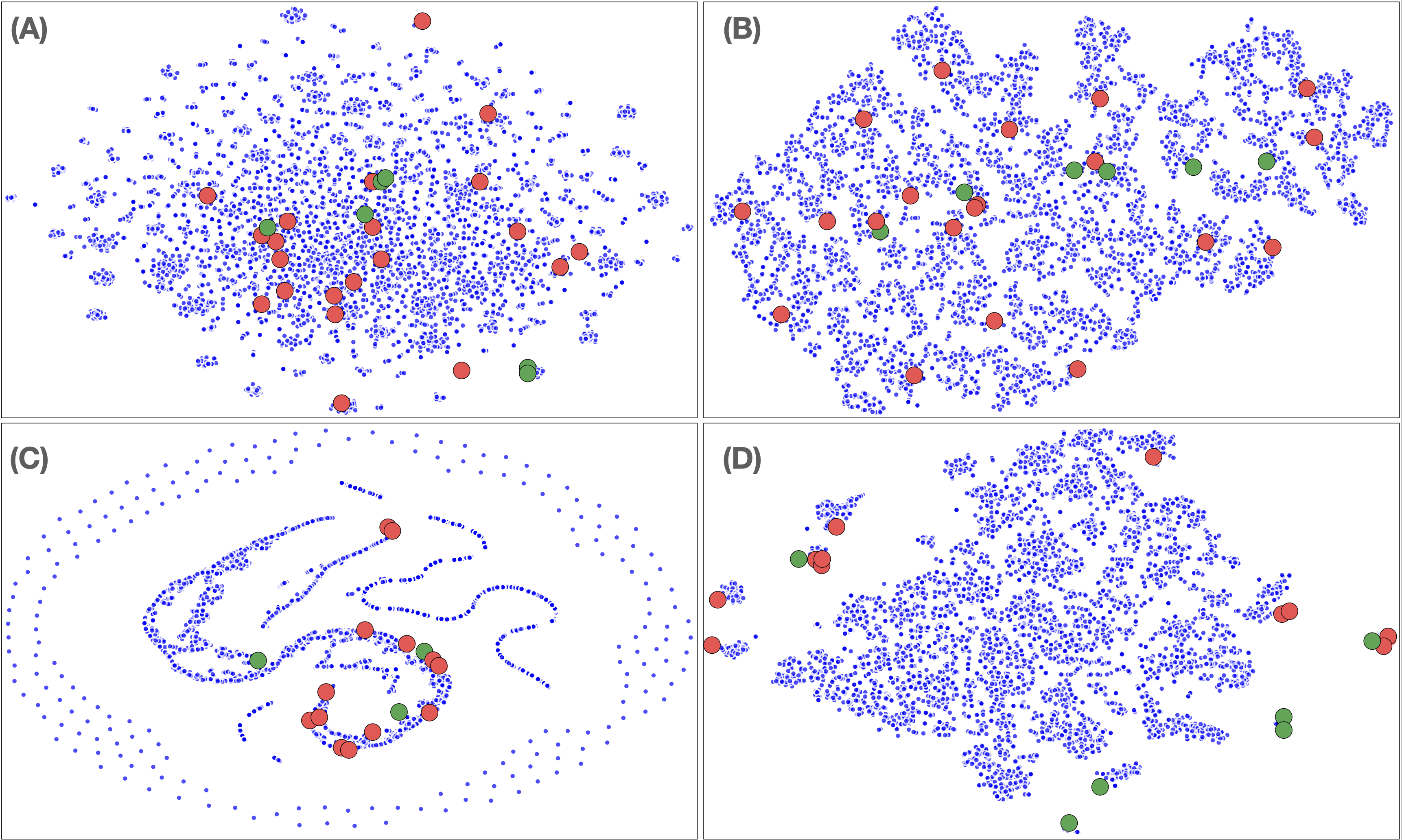}
    \caption{Visualizing edge embeddings for a 1000-node Barab\'{a}si-Albert (BA) graph. Shown in red are edges that PATHATTACK deleted, shown in green are edges that are part of $p^*$, with blue being all other edges. The graph was embedded into 32 dimensions using DGI. Node features used were (A) no node features (B) structural features (C) flow values (D) Personalized Page Rank (PPR) vectors. For the BA graph, using PPR vectors as node features (D) leads to the deleted edges (red) separating out from the rest of the edges.}
    \label{fig:ba_dgi}
\end{figure}

Figures \ref{fig:lattice_dgi}  and \ref{fig:ba_dgi} show the results of embedding %a 900 node (30 x 30) lattice network 
two different graph topologies, a lattice and a Barab\'{a}si-Albert graph, respectively, into 32 dimensions using DGI. Highlighted in red are edges deleted by PATHATTACK, in green are edges along the desired path $p^*$ and in blue are all the remaining edges. Each of the 4 sub plots were generated using different sets of node features as discussed above. Figure \ref{fig:lattice_dgi},\ref{fig:ba_dgi} (A) are generated using no node features; \ref{fig:lattice_dgi}, \ref{fig:ba_dgi} (B) use structural features, \ref{fig:lattice_dgi}, \ref{fig:ba_dgi} (C) use flow values are node features and \ref{fig:lattice_dgi}, \ref{fig:ba_dgi} (D) use personalized page rank (PPR) vectors for each node along $p^*$ as node features. 

%Observe that the separability of deleted edges (red) is significantly better when flow information is passed in (\ref{fig:lattice_dgi}(C)). 

We observe that not all graph embeddings are equally effective in highlighting the structure of the PATHATTACK solution edges. In particular, embeddings that use the flow information (for the lattice graph) and the $p^*$ related PPR scores (for the Barab\'{a}si-Albert graph) lead to a separation of the solution edges from the rest of the graph. These observations lead to the main intuition of our ML-aided optimization algorithm. We can use carefully engineered graph representations to identify a much smaller subgraph that contains the solution edges, effectively reducing the dimensionality of the optimization space and allowing us to accelerate the solution identification run-time.

%However, this behaviour is not seen across all types of networks. Figure \ref{fig:ba_dgi} repeats the same experiment for a 1000 node Barab\'{a}si-Albert network. As before, Figure \ref{fig:ba_dgi}(A), (B), (C) and (D) correspond to using no node features, structural features, flow information and personalized page rank respectively. Here we observe that using personalized page rank Figure \ref{fig:ba_dgi}(D) leads to embeddings where deleted edges separate out better. 

\subsection{Problem Size Reduction via GRASP}
\label{gat}

To learn the subgraph of interest, we use supervised variants of graph neural networks. More specifically, we use Graph Attention Networks (GAT) \cite{velickovic2018graph}. We pass in each individual set of features discussed above as node features, as well as the concatenation of all three sets of features, with the results of each discussed in Section \ref{results}. GRASP starts by picking nodes in the 95th percentile of the GAT predictions as the initial subgraph and passes the resultant graph into PATHATTACK. It checks if the resultant path is the same as $p^*$, if not, reduces the percentile threshold (thereby generating a larger subgraph) and repeats the process until the resultant path that PATHATTACK returns is $p^*$. 

\subsubsection{Experimental Setup}
We consider both synthetic and real graphs for training and testing the GRASP algorithm.
The synthetic graph topologies include the Lattice, Barab\'{a}si-Albert, Erd\H{o}s-R\'{e}nyi and Watts-Strogatz models. Each graph instance generated contained $n=1000$ nodes and average edge density of $n\log{n}$. Edge densities were chosen to keep the graphs sparse, a domain which the PATHATTACK algorithm finds difficult to solve. We trained a different graph attention model for each type of graph topology, using 25 instantiations, where the source and target nodes are chosen randomly each time, with $p^*$ being the $800$th shortest path. Table \ref{tab:param} shows the distribution of all the varying graph parameters. 
Real world graphs were however, chosen to span a wide range of edge densities, sizes and characteristics to demonstrate the scalability of GRASP. Details of the real graphs chosen are shown in Table \ref{tab:real}. The CPG graph has a grid like topology with AS and LBL graphs being computer networks with topologies akin to scale free graphs. The Wikispeedia graph is a dense graph with a large number of triangles. Weights and costs on all edges were set to 1 and each graph was trained for 1000 epochs, with an early stopping criteria of 250 steps. %Further details of experimental setup are available on our online repository. We select the 800th shortest path as $p^*$. 

For the testing setup, we generate 15 graphs of each topology type and report averaged results across these runs. For real graphs, we test on 15 randomly drawn triplets of source, target and $p^*$'s.  We also compare against a greedy baseline algorithm that finds the shortest path between source and target and deletes the edge with the minimum cost. This process is repeated till $p^*$ is the shortest path. We refer the interested reader to the paper by Miller et al. \cite{miller2022attacking} for more details on PATHATTACK and the baseline algorithm (amongst baselines discussed in \cite{miller2022attacking}, this one prioritises shorter run times).

\subsubsection{Hyperparameters}
The learnable model used consists of two graph attention layers with 16 and 32 attention heads respectively, followed by three dense fully connected layers with 32 nodes in each layer. Dropout layers with probability of dropping nodes of 60\% are used after each layer along with batch normalization. The overall model consists of ~275k parameters. The models were trained from scratch, with one GNN model per graph topology type. Interestingly, we notice that the model generalises better if trained for fewer epochs per graph instance with repeated passes over the instances. Concretely, given 100 BA graph instances, the model generalizes better when trained for 500 epochs per instance, with 10 passes over the 100 instances, rather than 1000 epochs with 5 passes over the same 100 instances.

\subsubsection{Thresholding}
The output of the learned model are continuous values between 0 and 1. In order to select a subset of nodes, GRASP selects nodes at a predefined threshold and passes the selected nodes into PATHATTACK. In practice, the initial threshold is set as the 95th percentile. If the shortest path returned by PATHATTACK is not $p^*$, GRASP reduces the threshold and selects a larger subset of nodes. We set the decrement size to 10. %, i.e., the second step would select all nodes at the 85th percentile, followed by 70th percentile etc.. }

The starting percentile and the step size have an impact on the run time, in terms of number of calls made to PATHATTACK versus size of the input problem given to PATHATTACK. However, after each step of the thresholding procedure, the partial solution found is carried over into the next step, i.e., the edges deleted in the first step remain deleted in the following steps, due to which, we see a speed up in the run time even in the case when the thresholding procedure goes all the way down to 0 and selects the entire graph.

\begin{table}
  \caption{Parameters ranges for the synthetic graphs. Each synthetic graph instance contained about $n=1000$ nodes and about $|E|\approx n\log n \approx 7000$ edges. The parameters $p$ stands for probability of edge creation for ER graphs, $m$ stands for number of edges to attach from a new node to existing nodes for BA graphs and $k$ is the number of nearest neighbors each node is joined to in a ring topology for WS graphs with $p_r$ being the probability of rewiring each edge.}
  \label{tab:param}
  \begin{tabular}{ccc}
    \toprule
    Graph & Parameter(s) &Range \\
    \midrule
    Erd\H{o}s-R\'{e}nyi (ER) &$p$ & 0.01-- 0.017 \\
    Barab\'{a}si-Albert (BA)& $m$ & 5--9 \\
    Watts-Strogatz (WS) & $k$, $p_r$ & 11--15, 0.02\\
  \bottomrule
\end{tabular}
\end{table}

\section{Results}
\label{results}

\begin{figure*}
    \centering
    \includegraphics[width=0.99\linewidth]{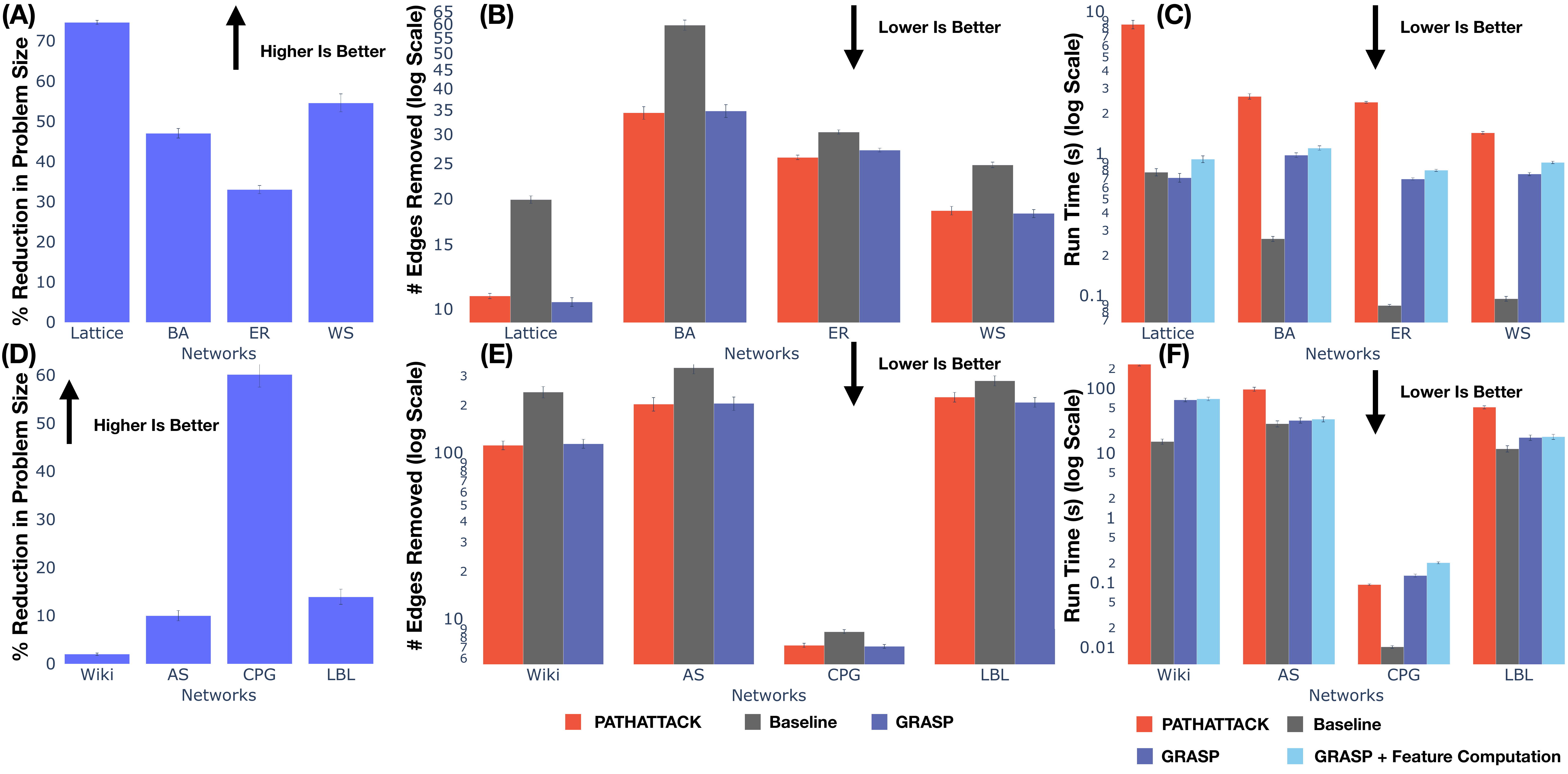}
    \caption{The first row of figures (i.e., A, B, and C) show results on synthetic graphs. The second row of figures (i.e., D, E, and F) show results on the real-world graphs listed in Table \ref{tab:real}. Figures (A) and (D) show the percentage reduction in input problem size with GRASP compared to the original input graph used by PATHATTACK and baseline. Figures (B) and (E) show the number of edges removed. Observe that GRASP and PATHATTACK remove fewer edges than the baseline. Figures (C) and (F) show that run times for GRASP are shorter across graphs when compared to PATHATTACK. (The exception is CPG, which is too small to benefit from the edge selection procedure.) GRASP achieves the same performance as PATHATTACK and typically runs twice as fast, while the baseline runs faster with significant additional cost in the number of edges removed.}%Baseline achieves the shortest run times, however it trades off solution quality (as measured by the number of edges removed).}
    \label{fig:results}
\end{figure*}

Figure \ref{fig:results} shows runtime performance across the different graph topologies. For this experiment, we considered PPR vectors as node features considering their success in learning using graph embeddings relative to the optimization task. Subplots (A), (B) and (C) show results for synthetic graphs with subplots (D), (E) and (F) showing results for real graphs. Across Figures \ref{fig:results}(C) and (F), we see that run times for GRASP is shorter than PATHATTACK, with the largest reduction being ~10x for Lattice graphs (C). We also show run times that include the time taken to compute the node features and show that it is still shorter than running PATHATTACK. Figures \ref{fig:results}(A) and (D) show the reduction in problem size across graphs. We see that reduction in problem sizes are largest for grid like graphs (Lattice and CPG), with reductions being smallest for ER graphs. Figures \ref{fig:results}(B) and (E) show the number of edges cut, where we see GRASP cutting similar number of edges as PATHATTACK, sometimes outperforming PATHATTACK. 

An interesting case is that of the CPG graph where the percentage of reduction is large, with the number of edges cut being fewer when using GRASP as compared to PATHATTACK, but the run time being larger. We posit the larger run times are due to the graph being small ($\sim$300 nodes), where it might make more sense to simply run the original PATHATTACK algorithm. 

\begin{table*}
  \caption{Real-world graphs used and their properties. For each graph, we list the average degree ($\langle k\rangle$), standard deviation of the degree ($\sigma_k$), average clustering coefficient ($\kappa$), transitivity ($\tau$), number of triangles ($\triangle$), and number of components ($\varphi$).}
  \label{tab:real}
  \begin{tabular}{ccccccccc}
    \toprule
   Graphs & Nodes & Edges &  $\langle k\rangle$ & $\sigma_k$ & $\kappa$ & $\tau$ & $\triangle$ & $\varphi$ \\
   \midrule
    Chilean Power Grid (\textbf{CPG}) \cite{kim2018depth} & 347 & 444 & 2.559 & 1.967 & 0.086 & 0.087 & 40 & 1 \\ 
    Lawrence Berkley Lab (\textbf{LBL}) \cite{lbldata} & 3,186 & 15,553 & 9.763 & 40.702 & 0.048 & 0.001 & 1821 & 10 \\
    Wikispeedia (\textbf{Wiki}) \cite{west2009wikispeedia} & 4,592 & 119,882 & 52.213 & 78.601 & 0.195 & 0.158 & 550,545 & 2 \\
    Autonomous System (\textbf{AS}) \cite{leskovec2005graphs} & 10,670 & 22,002 & 4.124 & 31.986 & 0.296 & 0.009 & 17,144 & 1 \\
  \bottomrule
\end{tabular}
\end{table*}

\begin{figure}[H]
    \centering
    \includegraphics[width=0.99\linewidth]{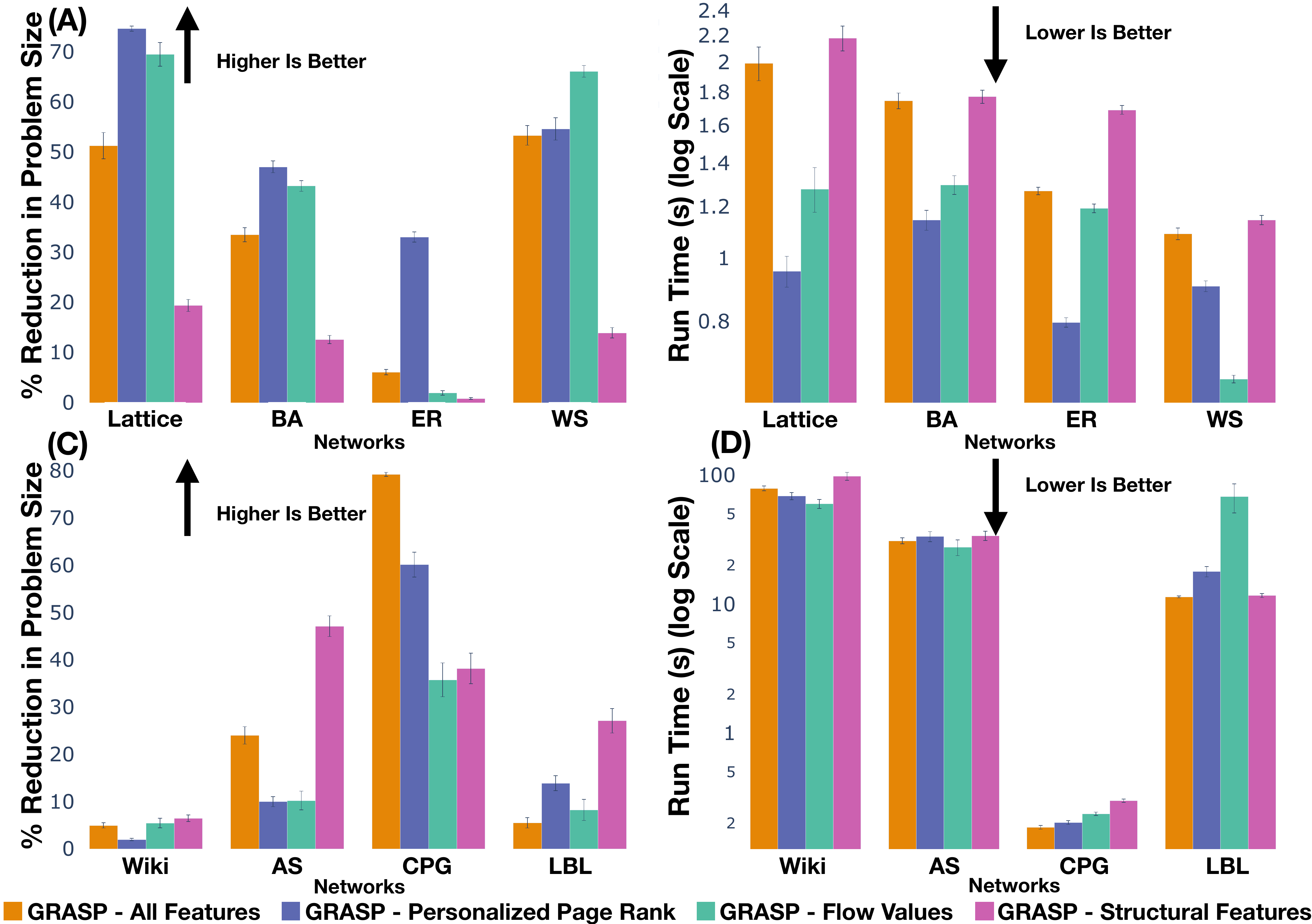}
    \caption{Figures (A) and (B) show results for synthetic graphs, with (C) and (D) showing results for real graphs from Table \ref{tab:real}. For synthetic graphs, using PPR as node features seems to have the highest reduction in problem size and lowest run times. However, for real graphs, using structural features seems to be the best choice. Problem size reduction is dependent on picking the right set of features, which in turn depends on the topology of the graph.}
    \label{fig:ablation}
\end{figure}

We compare the impact of the different sets of node features in Figure \ref{fig:ablation}. Results for synthetic graphs are shown in Figure \ref{fig:ablation} (A) and (B). We observe that for this set of synthetic topologies, using PPR only (red) yields the highest reduction in the problem size, as well as lowest run times. However, examining Figures \ref{fig:ablation} (C) and (D), we see that this is not the case for real world graphs. Here, we observe that structural features (purple) lead to higher reduction in problem size in 3 out of the 4 graphs tested. Interestingly, the concatenation of all features (blue) does not out perform any one set. Since we run the resultant subgraph through the PATHATTACK algorithm, we always ensure solution quality. However, picking the right set of node features has a large impact in the reduction of the problem size and the resultant run times, and depends on the underlying topology of the graph. 

\begin{figure}[H]
    \centering
    \includegraphics[width=0.99\linewidth]{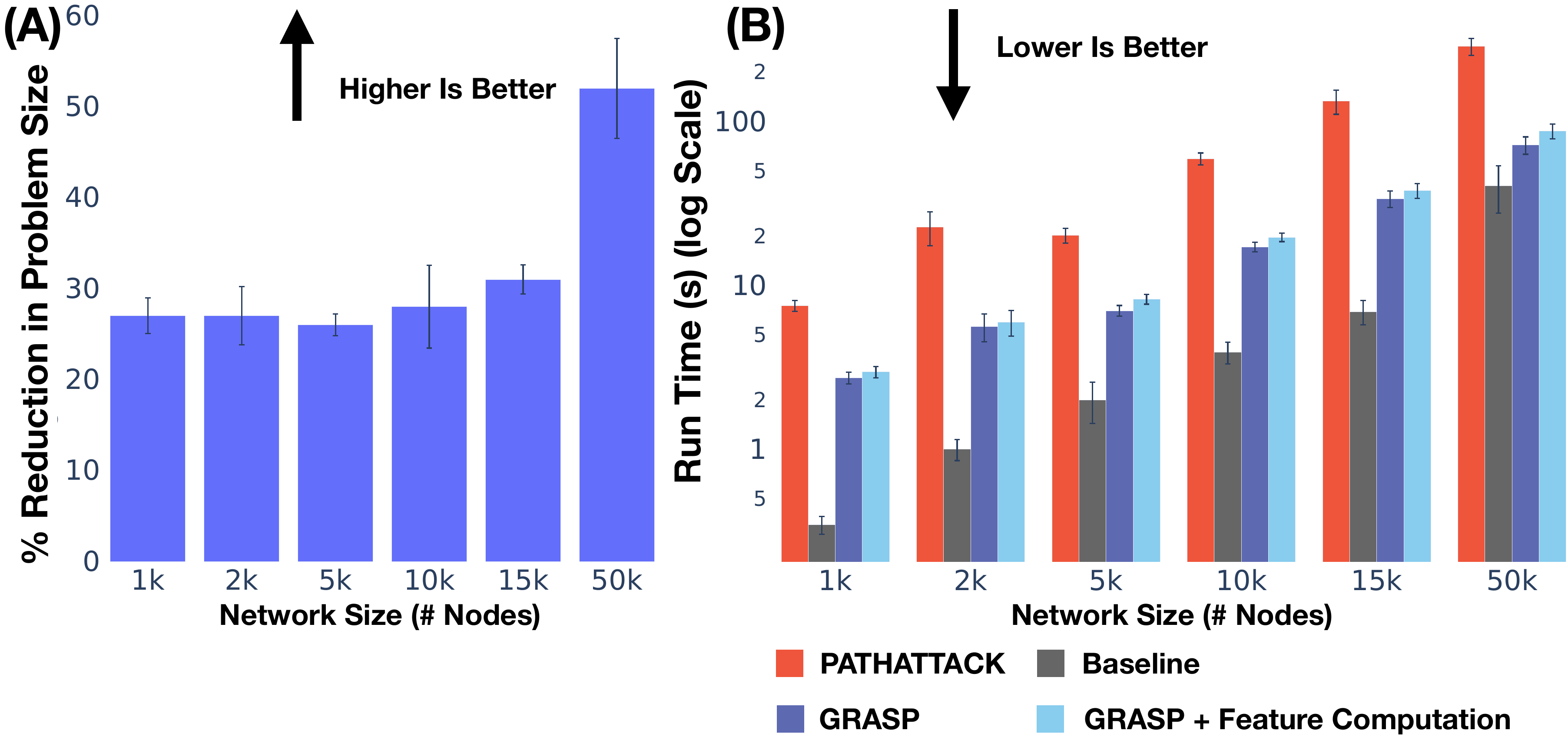}
    \caption{We demonstrate the scalability of GRASP trained on a BA graph with 1000 nodes and tested across BA graphs ($m = 7$) of varying sizes. Figure (A) shows reduction in problem size and (B) shows reduction in run times, where we see that GRASP scales well with input graph size.}
    \label{fig:scale}
\end{figure}

The experiments on real world graphs demonstrate GRASP across graphs with varying sizes, graph characteristics and topologies. We also highlight the scalability by running GRASP trained on a BA ($m = 7$) graph with 1000 nodes and tested on BA graphs of varying sizes in Figure \ref{fig:scale}. 

\section{Related Work}

ML has been used to approximate classical algorithms or combinatorial problems in various forms. The Neural Algorithmic Reasoning (NAR) framework proposed by Veli{\v{c}}kovi{\'c} et al. \cite{velivckovic2021neural} exploits the fact that graph neural networks have similar structure to classical dynamic programming problems to learn a variety of commonly used classical algorithms \cite{velivckovic2022clrs}. Numeroso et al. \cite{numeroso2023dual} take the NAR framework a step further and propose Dual Algorithmic Reasoning that aims to jointly optimize for the primal and dual of a given problem. Specifically, they look at training models to find solutions for both the min cut and the max flow at the same time and show that it leads to better solutions. 

In terms of NP-hard problems, Khalil et al. \cite{khalil2017learning} propose a reinforcement learning approach that utilizes graph embeddings to learn a greedy policy that incrementally builds a solution. Closely related to our method is work by Li et al. \cite{li2018combinatorial}, where a GCN is used as a heuristic function for a greedy search algorithm for the MIP solver. More specifically, the input graph is reduced to a smaller graph based on probability maps generated from the GCN that encode the likelihood of each node being in the optimal solution. Gasse et al. \cite{gasse2019exact} use a GNN to learn expensive to compute branch-and-bound heuristics for MIP solvers. The learned models are then used at run time to achieve speed ups while maintaining the quality of the solution. 

In this context, we can categorize the use of ML in combinatorial optimisation as belonging to either an end to end approach, or one where the learned model is used to aid an existing classical algorithm. We point the interested reader to the survey by Bengio et al. \cite{bengio2021machine} that contains a deeper dive into the different categorisations and their details. 

% We need to compare to this work: Learning Combinatorial Optimization Algorithms over Graphs, Dai et al.

% Combinatorial Optimization with Graph
% Convolutional Networks and Guided Tree Search, Li et al.

% Machine Learning for Combina-
% torial Optimization:
% a Methodological Tour dHorizon, Bengio et al. (check pages 18,19)

\section{Limitations}
\begin{itemize}

    \item GRASP can occasionally end up running the thresholding procedure until the entire graph has been selected. We do however notice that doing so still runs faster than PATHATTACK since the solution at each thresholding step is carried over into the next step. 
    \item For small graphs ($n \leq 500$), the computational overhead of GRASP turns out to be expensive and PATHATTACK runs faster (see Figure \ref{fig:results}(F) for an illustration).
\end{itemize}

\section{Conclusion and Future Work}
We consider the Force Path Cut problem, where an adversary's aim is to cut a minimum number of edges in order to force traffic from a pair of nodes through a specific path. This is an APX-hard problem. The PATHATTACK algorithm~\cite{miller2022attacking} approximately solves this problem by using a linear programming formulation of the Weighted Set Cover problem. We present GRASP, an ML-based optimization method that speeds up run time while maintaining the quality of solutions generated. We highlight the importance of various node features and how they impact run time and solutions generated. We plan to extend this work to speed up general set cover problems by (1) reducing the input problem space and (2) running the classical algorithms on the reduced input space.

%% The next two lines define the bibliography style to be used, and
%% the bibliography file.
\bibliographystyle{ACM-Reference-Format}
\bibliography{bib}

\end{document}